\pdfoutput=1

\documentclass[11pt]{article}

\usepackage[]{ACL2023}

\usepackage{times}
\usepackage{latexsym}
\usepackage{color}
\usepackage{amsmath}
\usepackage{adjustbox}
\usepackage{booktabs}
\usepackage{multirow}
\usepackage{makecell}
\usepackage{float}

\usepackage{xcolor}
\usepackage{tcolorbox}
\tcbuselibrary{most}
\usepackage{colortbl}
\usepackage{diagbox}
\usepackage{amsthm,amsmath,amssymb}
\usepackage{mathrsfs}

\definecolor{lightgray}{gray}{0.9}

\usepackage[T1]{fontenc}

\usepackage[utf8]{inputenc}

\usepackage{microtype}

\usepackage{inconsolata}
\usepackage{graphicx}
\usepackage{xspace}

\title{\textsc{TimeToM}: Temporal Space is the Key to Unlocking the Door of Large Language Models’ Theory-of-Mind}

\author{Guiyang Hou, Wenqi Zhang$^{\dagger}$, Yongliang Shen, Linjuan Wu, Weiming Lu$^{\dagger}$ \\
        College of Computer Science and Technology, Zhejiang University \\ \texttt{\{gyhou, zhangwenqi, luwm\}@zju.edu.cn}}

\begin{document}
\maketitle
\begin{abstract}
Theory of Mind (ToM)—the cognitive ability to reason about mental states of ourselves and others, is the foundation of social interaction. Although ToM comes naturally to humans, it poses a significant challenge to even the most advanced Large Language Models (LLMs). Due to the complex logical chains in ToM reasoning, especially in higher-order ToM questions, simply utilizing reasoning methods like Chain of Thought (CoT) will not improve the ToM capabilities of LLMs. We present \textsc{TimeToM}\xspace, which constructs a temporal space and uses it as the foundation to improve the ToM capabilities of LLMs in multiple scenarios. Specifically, within the temporal space, we construct Temporal Belief State Chain (TBSC) for each character and inspired by the cognition perspective of the social world model, we divide TBSC into self-world beliefs and social world beliefs, aligning with first-order ToM (first-order beliefs) and higher-order ToM (higher-order beliefs) questions, respectively. Moreover, we design a novel tool-belief solver that, by considering belief communication between characters in temporal space,  can transform a character's higher-order beliefs into another character's first-order beliefs under belief communication period. 
Experimental results indicate that \textsc{TimeToM}\xspace can dramatically improve the reasoning performance of LLMs on ToM questions while taking a big step towards coherent and robust ToM reasoning.
\end{abstract}

\section{Introduction}
\renewcommand{\thefootnote}{\fnsymbol{footnote}}
\footnotetext[2]{\;Corresponding author.}
\renewcommand{\thefootnote}{\arabic{footnote}}

Humans continually try to reason about other people's mental states and understand how it might impact their actions \cite{frith2003development}. This capability, known as Theory of Mind (ToM) \cite{premack1978does}, is crucial for social interactions. With Large Language Models (LLMs) playing a growing role in our lives, developing LLMs with ToM could be better at teaching us, learning from us, communicating with us, collaborating with us, and understanding us \cite{gandhi2021baby, gandhi2023understanding, rabinowitz2018machine, shu2021agent}.

\begin{figure*}
    \centering
    \includegraphics[width=0.99\textwidth]{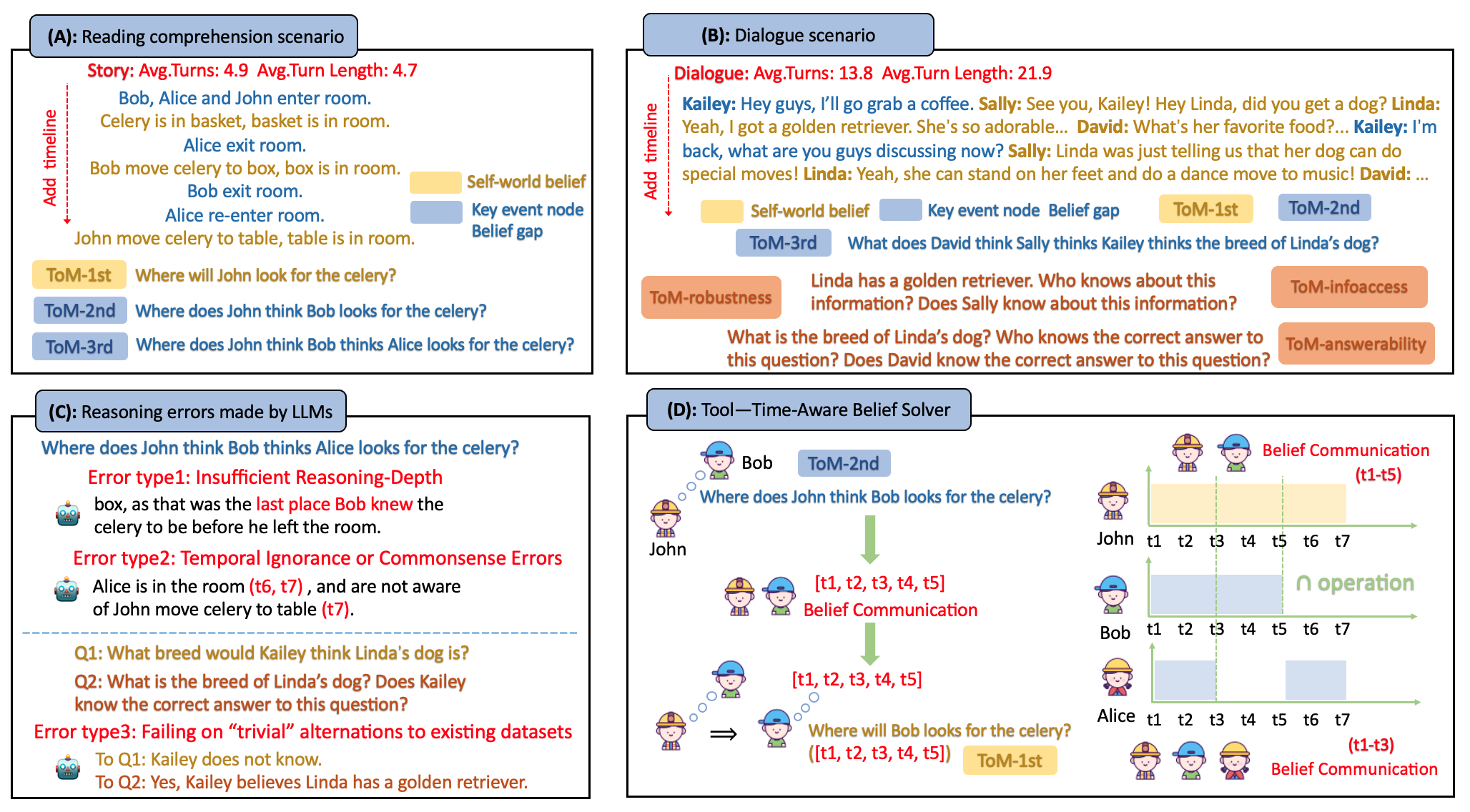}
    \caption{(A) and (B): The structure of story and dialogue, as well as ToM questions in reading comprehension and dialogue scenarios. (C): ToM reasoning errors made by LLMs. (D): Reasoning perspective of belief solver.}
    \label{fig:example1}
    \vspace{-3mm}
\end{figure*}

Although ToM often comes naturally to humans, LLMs often make various errors in ToM reasoning (Figure.\ref{fig:example1}C), such as ignoring the temporal order of events, generating outputs that violate commonsense, confusing the reasoning logic in higher-order ToM questions \cite{he2023hi} and failing on "trivial" alternations to existing datasets \cite{kim2023fantom, ullman2023large}. Recently, various reasoning strategies like Chain of Thought (CoT) \cite{wei2022chain} and Tree of Thought (ToT) \cite{yao2023tree} have improved the reasoning abilities of LLMs in some tasks. However, these strategies are not suitable for ToM reasoning \cite{ma2023towards}. Furthermore, \citet{wilf2023think} proposes the perspective-taking strategy to improve the ToM reasoning abilities of LLMs, but this strategy is not suitable for higher-order ToM reasoning. 
Currently, there is still a lack of effective reasoning strategies to improve the performance and robustness of LLMs in ToM reasoning tasks.

In this paper, we introduce \textsc{TimeToM}\xspace, which initially constructs temporal space by adding timeline to the stories or dialogues. Within the temporal space, we construct Temporal Belief State Chain (TBSC) for each character based on the events they are aware of on the timeline. Meanwhile, inspired by a principle of modern cognitive science \cite{doi:10.1126/science.adm8175, yue2022world}, which posits that humans construct abstract models of the social world and their self-world in their minds, we split the beliefs in TBSC into self-world beliefs and social world beliefs. We use self-world beliefs to answer first-order ToM questions and incorporate social world beliefs when answering higher-order ToM questions. 

The reasoning difficulty of ToM questions significantly increases as the order of questions rises \cite{sclar2023minding}, and currently, there is no effective reasoning strategy for solving higher-order (reasoning depth $m \geq 2 $) ToM questions. We consider that the key to higher-order ToM reasoning lies in capturing the belief communication between characters. We design a novel tool—belief solver, which first parses each character's perceptible time set based on their TBSC and then calculates the intersections of the time set of different characters to determine at which times they achieve belief communication. Furthermore, as illustrated in Figure.\ref{fig:example1}D, since John's understanding of Bob's belief only occurs during the belief communication period, the higher-order ToM question of John's estimation of Bob's beliefs can be transformed into a first-order ToM question of what Bob's belief state is like during the belief communication period. In reasoning about higher-order ToM questions, LLMs generate an initial reasoning process based on the character's TBSC, and the belief solver transforms the higher-order ToM questions into first-order ToM questions under belief communication period, which serves as feedback to inspire LLMs to refine their initial reasoning process on higher-order ToM questions.

Experimental results Experimental results on the ToMI \cite{le2019revisiting}, BigToM \cite{gandhi2023understanding}, and FanToM \cite{kim2023fantom} benchmarks indicate that \textsc{TimeToM}\xspace dramatically improves the reasoning performance of LLMs on ToM questions in multiple scenarios (Figure.\ref{fig:example1}A and Figure.\ref{fig:example1}B), while taking a big step towards coherent and robust ToM reasoning. It's noteworthy that \textsc{TimeToM}\xspace is well-suited for higher-order ToM questions, demonstrating good performance in third-order ToM questions. Furthermore, \textsc{TimeToM}\xspace can be applied to situations involving agent communication which is commonly occurring in the real-world.

The main contributions of this work are as follows: (1) We construct temporal space and use it as the foundation. (2) Within the temporal space, we construct TBSC, which is a comprehensive representation of a character's beliefs, including the temporal evolution of thing states and clear temporal cognition of key social events that create belief gaps. From the social world model's cognitive perspective, TBSC is split into self-world and social world beliefs, aligning with first-order ToM (first-order beliefs) and higher-order ToM (higher-order beliefs) questions respectively. (3) Within the temporal space, we design a novel tool—belief solver, which serves as feedback to inspire LLMs to refine their initial reasoning process in higher-order ToM questions. (4) Temporal space is the key to unlocking the door of LLMs' Theory of Mind. Extensive experimental results indicate that \textsc{TimeToM}\xspace dramatically improves the reasoning performance and robustness of LLMs on ToM questions\footnote{Code will available at \url{
https://github.com/gyhou123/TimeToM}.}.

\section{Background and Related Work}
\label{sec:appendix0}

\paragraph{Existing ToM Benchmarks}
Previous evaluations for the ToM of LLMs are primarily focused on testing models using situation descriptions (i.e., narratives)  \cite{nematzadeh2018evaluating, le2019revisiting, sapNeuralTheoryofMindLimits2022, shapira2023clever}, also referred as reading comprehension scenarios. Recently, considering ToM capabilities play an even more important role in understanding dynamic social interactions, \citet{kim2023fantom} introduce FanToM, which tests models using interactive dialogues, also referred as dialogue scenarios.

\paragraph{LLMs Lack of ToM Capabilities}

Several studies \cite{gandhi2023understanding, sapNeuralTheoryofMindLimits2022,kim2023fantom,wilf2023think,ullman2023large} have shown that LLMs have poor reasoning performance and robustness on ToM tasks in a zero-shot setting, even with the current state-of-the-art GPT-4 \cite{achiam2023gpt} model. With LLMs becoming increasingly integrated into our everyday lives, developing LLMs with ToM is very necessary.

\paragraph{Enhancing LLMs Reasoning Capabilities} Recent prompt-based methods enhance the reasoning abilities of LLMs by guiding them to produce intermediate reasoning steps. For example, CoT \cite{wei2022chain} guides LLMs to generate step-by-step derivations before producing the final answer, LtM \cite{zhou2022least} decomposes a target question into a series of subquestions, ToT \cite{yao2023tree} using tree-structured search to find better reasoning chains, Self-Refine \cite{madaan2023self} and Reflexion \cite{shinn2023reflexion} adds self-verification steps for rectifying reasoning errors. RAP \cite{hao2023reasoning, hu2023language} repurposes an LLM as a world model by prompting the LLM to predict the next state $s_{t+1}$ of reasoning after applying a reasoning step $a_{t}$ to the current state $s_{t}$. Apart from prompt-based methods, Declarative \cite{he2023solving} uses external symbolic solver to solve the equations in reasoning steps. MAF \cite{nathani2023maf} uses multiple external tools such as calculator, programming syntax to generate feedback to refine initial reasoning output. 

Although many methods have been introduced to enhance the reasoning ability of LLMs, they are not suitable for ToM reasoning. \citet{wilf2023think} adopts the perspective-taking strategy to enhance the ToM reasoning abilities of LLMs, but this strategy falls short in addressing higher-order ToM reasoning. \citet{sclar2023minding}\footnote{This method works by explicitly memorizing beliefs of each character, rather than utilizing LLMs for ToM reasoning} tracks each entity's beliefs and their estimation of other entities’ beliefs, through graphical representations. However, it requires a substantial amount of external memory as well as being difficult to apply to context-rich ToM scenarios.  
Furthermore, there are currently no tools specifically dedicated to ToM reasoning.

\section{\textsc{TimeToM}\xspace Overview}
Figure.\ref{fig:example1}C illustrates the errors LLMs made in ToM reasoning, often ignoring the temporal order of events and confusing the reasoning logic in higher-order ToM questions. Concurrently, explicitly representing the timeline not only allows LLM to have a clearer temporal understanding of the events in the story and dialogue, but we also observe the association between higher-order questions (higher-order beliefs) and first-order questions (first-order beliefs) on the timeline. Building upon this insight, we introduce \textsc{TimeToM}\xspace, improving the ToM capabilities of LLMs in interactive dialogue and reading comprehension scenarios. The overall procedure of \textsc{TimeToM}\xspace is shown in Figure \ref{fig:method1}.


\subsection{Constructing Temporal Space}
\label{subsec:timeline}
\begin{figure*}
    \centering
    \includegraphics[width=1\textwidth]{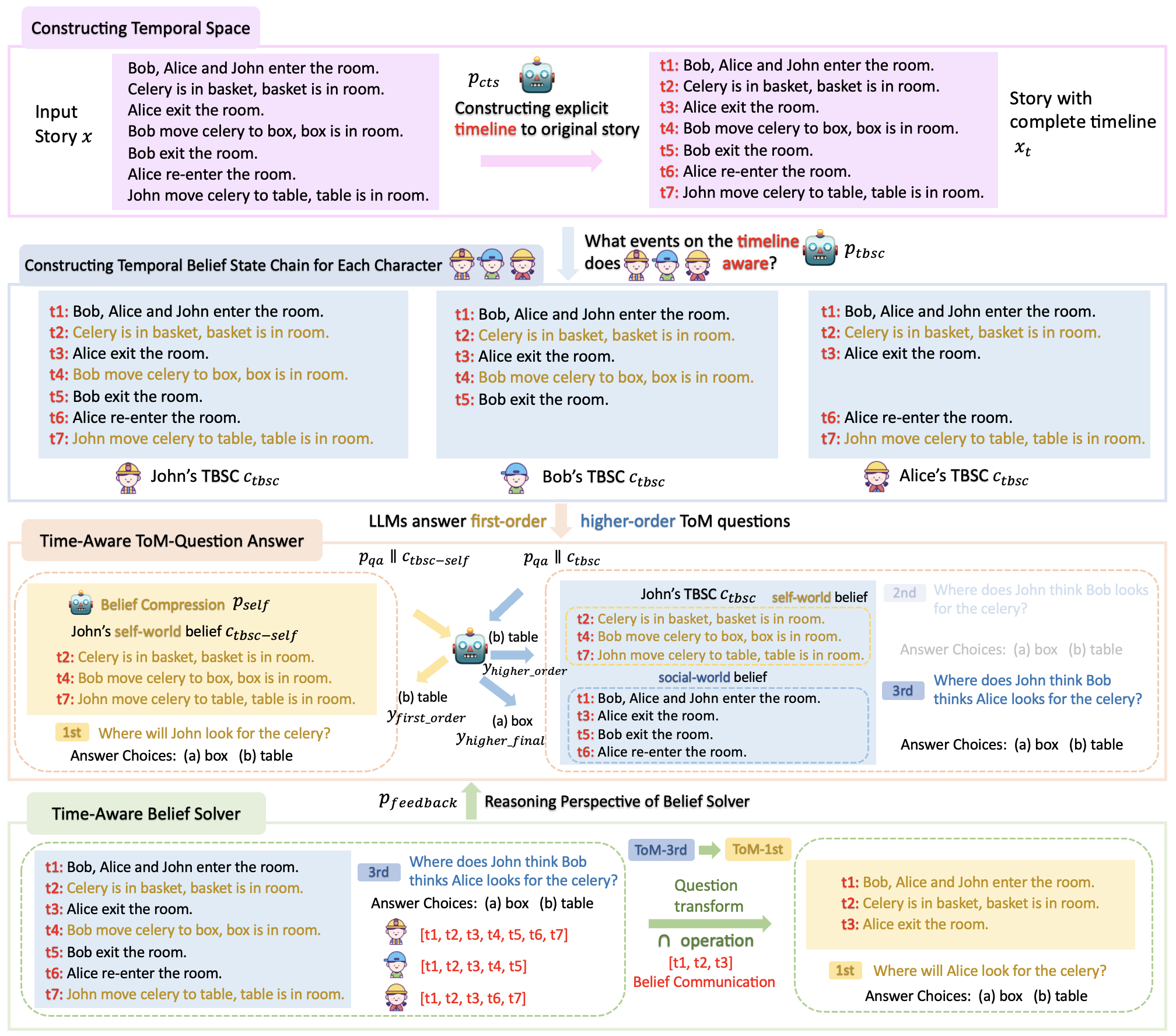}
    \caption{Pipeline overview of \textsc{TimeToM}\xspace, which constructs a temporal space and uses it as the foundation to improve the ToM capabilities of LLMs. \textsc{TimeToM}\xspace does not require training, it works in a zero-shot setting.}
    \label{fig:method1}
    \vspace{-3mm}
\end{figure*}

In reading comprehension scenarios, each sentence in the story corresponds to a specific time point. Similarly, in interactive dialogue scenarios, each utterance in the dialogue corresponds to a specific time point. Illustrating with the case of reading comprehension scenarios, given the input story $x$: {\textit{ Sentence1, Sentence2,..., SentenceN}}, prompt $p_{cts}$, and model $\mathcal{M}$, \textsc{TimeToM}\xspace adds a complete timeline for input story $x$ to form $x_t$:
\begin{equation}
    x_t = \mathcal{M}(p_{cts}||x).
\end{equation}
For example, as illustrated in Figure \ref{fig:method1}, the model explicitly adds time points before each sentence for the given input story. Here, $||$ denotes concatation and $p_{cts}$ is shown in Appendix \ref{appendix2}.


\renewcommand{\dblfloatpagefraction}{.8}

\subsection{Constructing Temporal Belief State Chain for Each Character}
\label{subsec:tbsc}
ToM questions focus on the beliefs of characters, including their own beliefs as well as their estimations of others' beliefs. Given story $x_{t}$ within temporal space, prompt $p_{tbsc}$ and model $\mathcal{M}$, \textsc{TimeToM}\xspace construct TBSC $c_{tbsc}$ for each character, based on the events they are aware of on the timeline:
\begin{equation}
    c_{tbsc} = \mathcal{M}(p_{tbsc}||x_{t})
\end{equation}
For example, as illustrated in  Figure \ref{fig:method1}, Alice is aware of events between {\color{red} \textbf{t1}} to {\color{red} \textbf{t3}}, but since she leaves the room at {\color{red} \textbf{t3}} and re-enters the room at {\color{red} \textbf{t6}}, he cannot aware of events between {\color{red} \textbf{t4}} to {\color{red} \textbf{t5}}. TBSC is a comprehensive representation of a character's beliefs, which includes the temporal evolution of object states, such as celery is in the basket at {\color{red} \textbf{t2}}, in the box at {\color{red} \textbf{t4}}, and on the table at {\color{red} \textbf{t7}} as well as key social events that create belief gaps with clear temporal logic, such as Alice exits the room at {\color{red} \textbf{t3}}, Bob exits the room at {\color{red} \textbf{t5}}, and Alice re-enters the room at {\color{red} \textbf{t6}}. Here, $||$ denotes concatation and $p_{tbsc}$ is shown in Appendix \ref{appendix2}.
\subsection{Time-Aware ToM-Question Answer from Social World Model Perspective}
\label{subsec:qa}
\citet{doi:10.1126/science.adm8175} posits that humans construct abstract models of the social world and their self-world in their minds. Inspired by this cognitive perspective, we divide the belief in TBSC into self-world belief and social world belief. We consider self-world belief as the perception of the state and information of things around oneself, while social world belief is the perception of other characters' actions that may lead to a belief gap\footnote{Considering self-world belief and social world belief in this way aligns well with first-order and higher-order ToM questions, making it very suitable for ToM scenarios.}. Given character's TBSC $c_{tbsc}$, prompt $p_{self}$ and model $\mathcal{M}$, \textsc{TimeToM}\xspace generates self-world belief $c_{tbsc-self}$ via belief compression, i.e., focusing on events in TBSC about the states and information of things:
\begin{equation}
    c_{tbsc-self} = \mathcal{M}(p_{self}||c_{tbsc}).
\end{equation}
Given the character's self-world belief $c_{tbsc-self}$ and the comprehensive belief $c_{tbsc}$ after incorporating social world beliefs, prompt $p_{qa}$, and model $\mathcal{M}$, \textsc{TimeToM}\xspace use self-world belief $c_{tbsc-self}$ to answer first-order ToM questions and comprehensive belief $c_{tbsc}$ when answering higher-order ToM questions:
\begin{align}
    \begin{split}
    y_{first\_order} &= \mathcal{M}(p_{qa}||c_{tbsc-self})\\
    y_{higher\_order} &= \mathcal{M}(p_{qa}||c_{tbsc}).
    \end{split}
\end{align}
For example, as illustrated in  Figure \ref{fig:method1}, John uses only self-world belief (celery is in the basket at {\color{red} \textbf{t2}}, in the box at {\color{red} \textbf{t4}}, and on the table at {\color{red} \textbf{t7}}) to answer the first-order ToM question "\textit{Where will John look for the celery?}" and incorporates social world belief (Alice exits the room at {\color{red} \textbf{t3}}, Bob exits the room at {\color{red} \textbf{t5}}, and Alice re-enters the room at {\color{red} \textbf{t6}}) when answering higher-order ToM questions "\textit{where does John think Bob looks for the celery?}". Here, $||$ denotes concatation, $p_{self}$ and $p_{qa}$ is shown in Appendix \ref{appendix2}.

\subsection{Time-Aware Belief Solver}
We achieve a more comprehensive and clearer representation of characters' beliefs by establishing a TBSC for each character, which improve the performance of LLMs in answering first-order and higher-order ToM questions. However, as the order of ToM questions increases, the depth of reasoning required becomes deeper, relying solely on the character's TBSC still leads to logical errors in reasoning. Through in-depth consideration of belief communication between characters, we design an external tool—belief solver, which provides a novel reasoning perspective that effectively reduces the depth of reasoning, and serves as a feedback to inspire LLMs to refine their initial reasoning process on higher-order ToM questions. 
\label{subsec:bs}
\paragraph{Time Set Parsing} We parse each character's perceptible time set based on their TBSC. For example, as illustrated in  Figure \ref{fig:method1}, the set of times that John, Bob, and Alice can perceive are as follows:
\begin{align}
    \begin{split}
    T_{John} &= [t_{1}, t_{2}, t_{3}, t_{4}, t_{5}, t_{6}, t_{7}] \\
    T_{Bob} &= [t_{1}, t_{2}, t_{3}, t_{4}, t_{5}] \\
    T_{Alice} &= [t_{1}, t_{2}, t_{3}, t_{6}, t_{7}]
    \end{split}
\end{align}

\paragraph{Belief Communication between Different Characters}
To determine at which times belief communication occurs between different characters, we perform intersection operations on the sets of times perceived by each character, as parsed in the previous step:
\begin{align}
    \begin{split}
    BC_{John,Bob} &= T_{John} \cap T_{Bob}\\
    BC_{John,Bob,Alice} &= T_{John} \cap T_{Bob} \cap T_{Alice} 
    \end{split}
\end{align}
where $BC_{John,Bob} = [t_{1},t_{2},t_{3},t_{4},t_{5}]$ represents the set of times for belief communication between John and Bob, the same applies to $BC_{John,Bob,Alice} = [t_{1}, t_{2}, t_{3}]$.
\paragraph{Transforming Higher-order ToM problems into First-order ToM problems} Consider second-order ToM question \textit{"Where does John think Bob looks for the celery?"}, since John's understanding of Bob's belief only occurs during  $BC_{John,Bob}$ period, this question can be transformed into a first-order ToM question \textit{"Where does Bob look for the celery?"} under $BC_{John,Bob}$. It is worth noting that this transformation is also applicable to third-order and higher-order ToM problems.

\definecolor{c1}{HTML}{AEC670}

{\renewcommand{\arraystretch}{1}
    \begin{table*}[t!] \begin{center}
    \begin{adjustbox}{width=\linewidth}
    \begin{tabular}{lcccccc}
            \toprule
            \multirow{2}{*}{Model}  
            & \multicolumn{4}{c}{\makecell{ToMI}}  
            & \multicolumn{2}{c}{\makecell{BigTOM}}\\
            \cmidrule(r{0.3em}){2-5} \cmidrule(r{0.3em}){6-7}  
            & Overall     & False-Belief  & First-Order   & Second-Order    & Overall      & False-Belief \\
            \midrule
            \multicolumn{7}{l}{\cellcolor{lightgray}0-Shot} \\
            Llama2-7b-chat & 44.50     & 28.25     &  39.00  &  40.00  &   52.50& 53.50\\
            Llama2-13b-chat &  51.00  &   39.25    &   54.75  &  34.75  &  55.25 & 46.50\\
            GPT-3.5-turbo &  68.60  &    67.25 &  68.75   &  52.75  &    78.50& 69.50   \\
            GPT-4 & 66.50   &   25.50 &  50.75   &  65.50  &  97.50  &  99.00  \\

            \multicolumn{7}{l}{\cellcolor{lightgray}0-Shot-CoT} \\
            Llama2-7b-chat & 43.70   &  24.00   &   45.00  & 37.75 & 50.50  & 39.50   \\
            Llama2-13b-chat &  45.00  &  16.50   &   43.00   &  37.00  & 57.25  & 52.50 \\
            GPT-3.5-turbo &  64.10  &  34.00  &  58.50   &  53.00  &  80.75  & 71.50   \\
            GPT-4 &  74.40  &  74.25 & 73.75   &  62.25  &  97.75  &  99.00  \\

            \multicolumn{7}{l}{\cellcolor{lightgray}\textsc{SimToM}\xspace} \\
            Llama2-7b-chat &  48.10  &   40.00 &   47.25  &  39.25  & 56.25   &  75.00  \\
            Llama2-13b-chat &  61.10  &   35.50 &  53.75   &  53.75  &  57.75  &  62.50  \\
            GPT-3.5-turbo &  72.80  &   81.00 &   74.75   &  57.25  &  84.00  &   78.00 \\
            GPT-4 & 87.80   &   87.75 &   93.75  &  75.75  & 96.00   &  98.00  \\

            \multicolumn{7}{l}{\cellcolor{lightgray}\textbf{\textsc{TimeToM}\xspace}} \\
            \multirow{2}{*}{Llama2-7b-chat} &  64.30  &   47.25 &    56.50  &  57.75  &  68.75  & 84.50   \\ & \textbf{(+19.80,+16.20)} &  \textbf{(+19.00, +7.25)} & \textbf{(+17.50, +9.25)} & \textbf{(+17.75, +18.50)} & \textbf{(+16.25, +12.50)} & \textbf{(+31.00, +9.50)}\\
            \multirow{2}{*}{Llama2-13b-chat} &  67.20     & 44.75    &  61.25   &  57.00  &  77.75  & 89.50   \\ 
            & \textbf{(+16.20, +6.10)} &   \textbf{(+5.50, +9.25)} &  \textbf{(+6.50, +7.50)} & \textbf{(+22.25, +3.25)} & \textbf{(+22.50, +20.00)} & \textbf{(+43.00, +27.00)}\\
            \multirow{2}{*}{GPT-3.5-turbo} & 80.80    &  82.00   &  80.50   &  71.50  &  93.75  &  96.00  \\
            & \textbf{(+12.20, +8.00)} &  \textbf{(+14.75, +1.00)} &  \textbf{(+11.75, +5.75)} & \textbf{(+18.75, +14.25)} & \textbf{(+15.25, +9.75)} & \textbf{(+26.50, +18.00)}\\
            \multirow{2}{*}{GPT-4} &  96.00  &   98.75 &   95.50   &  94.50  &  97.00  &  99.00  \\
            & \textbf{(+29.50, +8.20)} & \textbf{(+73.25, +11.00)} & \textbf{(+44.75, +1.75)} & \textbf{(+29.00, +18.75)} & \textbf{(-0.50, +1.00)} & \textbf{(+0.00, +1.00)}\\
            \midrule
            Avg & \textbf{(+19.43, +9.63)} &  \textbf{(+28.13, +7.13)} & \textbf{(+20.13, +6.06)} & \textbf{(+21.94, +13.69)} & \textbf{(+13.38, +10.81)} & \textbf{(+25.13, +13.88)}\\
            \bottomrule
    \end{tabular}
    \end{adjustbox}
    \caption{
          \textsc{TimeToM}\xspace results on ToMI across False-Belief, First-Order, Second-Order, and All question types. Since the BigToM benchmark contains only First-Order ToM questions, we only report results across False-Belief and All question types. We present \textbf{absolute accuracy difference} between \textsc{TimeToM}\xspace and the baselines (0-shot and \textsc{SimToM}\xspace). Results for True-Belief and Mem-Real question types can be found in Appendix \ref{appendix3}.
    }
    \vspace{-10pt}
    \label{tab:1}
\end{center}\end{table*}}

\paragraph{Inspiring LLMs to Reason on Higher-order ToM Questions}
%
Through in-depth consideration of belief communication between characters, we observe that higher-order ToM questions can be transformed into first-order ToM questions under belief communication periods. Given this reasoning process as feedback $p_{feedback}$, LLM's initial reasoning process $y_{higher\_order}$, and model $\mathcal{M}$, \textsc{TimeToM}\xspace lets the LLM reason again:
\begin{equation}
    y_{higher\_final} = \mathcal{M}(p_{feedback}||y_{higher\_order}).
\end{equation}
We hope that through this method, the LLM can pay attention to the belief communication between characters as well as the connection between higher-order beliefs and first-order beliefs to refine their initial reasoning outputs on higher-order ToM questions. Here, $||$ denotes concatation, $p_{feedback}$ is shown in Appendix \ref{appendix2}.

\section{Experiments}
\subsection{Settings}
\paragraph{Benchmarks} We evaluate \textsc{TimeToM}\xspace within reading comprehension and interactive dialogue scenarios, using ToMI, BigToM, and FanToM benchmarks. Compared to stories in reading comprehension scenarios, dialogues are more aligned with real-world scenarios requiring ToM reasoning. Furthermore, dialogues in FanToM are significantly longer, with a larger number of subtopics and characters per dialogue. This poses a greater challenge for LLMs, as it demands the LLMs’ capability to comprehend the complete dialogue utterance and reason about each character’s beliefs. Detailed explanations for story (or dialogue) structure, each type of question in ToMI, BigToM, and FanToM benchmarks, and evaluation metrics can be found in Appendix \ref{sec:appendix1}.

\paragraph{Baselines} We employ five widely utilized LLMs: three open source – Llama2-7b, 13b, and 70b chat ~\citep{touvron2023llama} – and two closed source: GPT-3.5-Turbo-0613 and GPT-4-0613 to evaluate \textsc{TimeToM}\xspace. To highlight the effectiveness of \textsc{TimeToM}\xspace,  we evaluate LLMs 0-shot on ToM benchmarks with and without our \textsc{TimeToM}\xspace. Additionally, we compare \textsc{TimeToM}\xspace with the CoT \cite{wei2022chain} and SimToM \cite{wilf2023think}, where SimToM is a recently proposed prompting framework specifically designed to improve the reasoning performance of LLMs on ToM questions, achieving state-of-the-art results. To make a fair comparison, we uniformly set the temperature to 0 (GPT-series models) or 0.3 (Llama2-series models) and top\_p to 0.95 for all experiments. We reproduce our \textsc{TimeToM}\xspace prompts in Appendix \ref{Our prompt}.




\subsection{Main Results}
In Tables \ref{tab:1} and \ref{tab:2}, we report the reasoning performance of LLMs for ToM questions in reading comprehension scenarios and dialogue scenarios. 
\paragraph{Substantial Improvement across Different LLMs and Scenarios}
From widely utilized commercial LLMs (GPT-series) to open-source models (Llama2-series), and from reading comprehension to interactive dialogue scenarios, \textsc{TimeToM}\xspace leads to substantial performance improvement. Specifically, in the reading comprehension scenario, we achieve an average absolute accuracy improvement of \textbf{+19.43\%} and \textbf{+9.63\%}, as well as \textbf{+13.38\%} and \textbf{+10.81\%} over the 0-shot and \textsc{SimToM}\xspace baselines for the ToMI and BigTOM benchmark, respectively. A larger improvement is observed in the interactive dialogue scenario, where \textsc{TimeToM}\xspace achieves an average absolute accuracy improvement of \textbf{+44.7\%} and \textbf{+13.6\%} over the 0-shot and \textsc{SimToM}\xspace baselines for the FanTOM benchmark. 

\begin{figure}[H]
    \centering
    \includegraphics[width=0.30\textwidth]{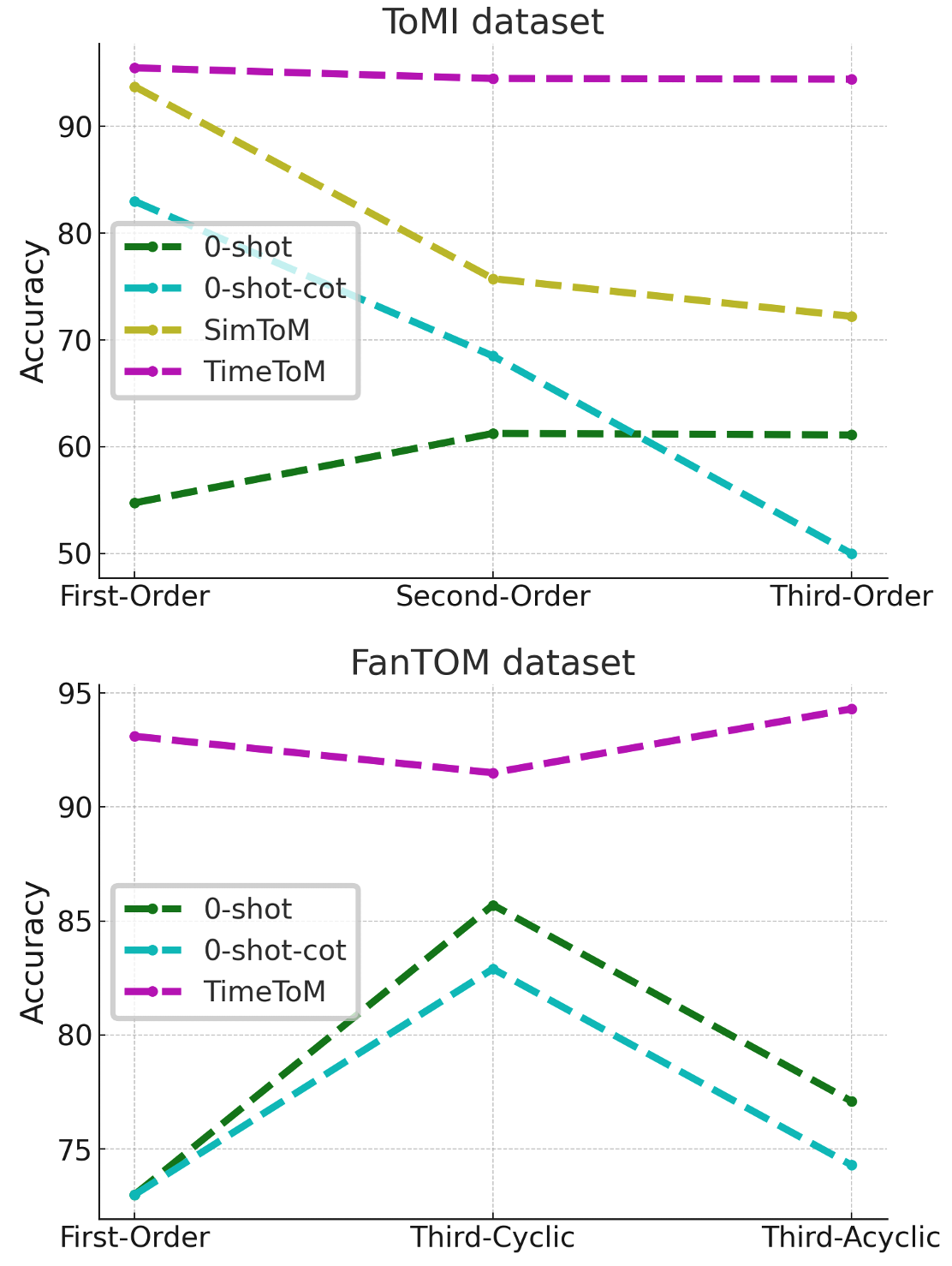}
    \caption{Performance comparison of \textsc{TimeToM}\xspace and baselines on first-order and higher-order ToM questions.}
    \label{fig:ana3}
    \vspace{-3mm}
\end{figure}

\paragraph{Well-suited for Higher-order ToM Reasoning} 
On the challenging higher-order ToM questions, \textsc{TimeToM}\xspace yields \textbf{+29.00\%} and \textbf{+18.75\%} absolute accuracy improvement over the 0-shot and \textsc{SimToM}\xspace GPT-4 baselines, as well as \textbf{+16.8\%} and \textbf{+17.9\%} absolute accuracy improvement over the 0-shot and 0-shot-CoT GPT-4 baselines for the ToMI and FanTOM benchmark, respectively. An equally impressive result is observed across other model types. Furthermore, based on the GPT-4 model, we test the performance of baselines and \textsc{TimeToM}\xspace on the third-order ToM problems of ToMI-Extend\footnote{\citet{sclar2023minding} construct third-order ToM questions by making simple modifications to the story structure of the original ToMI benchmark} benchmark. As shown in Figure \ref{fig:ana3}, compared with \textsc{SimToM}\xspace and 0-shot-CoT, the performance of \textsc{TimeToM}\xspace does not degrade as the order of the ToM question increases, indicating its suitability for higher-order ToM reasoning. Meanwhile, \textsc{TimeToM}\xspace exhibits the most outstanding performance on both first-order and higher-order ToM questions.

{\renewcommand{\arraystretch}{1}
    \begin{table*}[t!] \begin{center}
    \begin{adjustbox}{width=\linewidth}
    \begin{tabular}{lccccccccccc}
            \toprule
            \multirow{2}{*}{Model}  
            & \multicolumn{1}{c}{\multirow{2}{*}{\makecell{{\textcolor[RGB]{0,201,87} {\textbf{\textsc{All*}}}}\\Question\\Types}}}
            & \multicolumn{4}{c}{\makecell{{Belief}\\{Questions}}}  
            & \multicolumn{1}{c}{\multirow{2}{*}{\makecell{Answerability\\Questions\\{\textcolor[RGB]{0,201,87} {\textbf{All}}}}}}
            &\multicolumn{1}{c}{\multirow{2}{*}{\makecell{infoaccess\\Questions\\{\textcolor[RGB]{0,201,87} {\textbf{All}}}}}}\\
            \cmidrule(r{0.3em}){3-6} 
            &     & Overall     & First-order   & Third-acyc   & Third-cyc   &    &      \\
            \midrule
            \multicolumn{8}{l}{\cellcolor{lightgray}0-Shot} \\
            Llama2-70b-chat & 0.0 & 6.5 & 8.7 & 0.0 & 5.7 & 4.3 &  8.7 \\
            GPT-4 & 8.7 & 76.2 & 73.0 & 77.1 & 85.7 & 23.5 & 23.5 \\
            \multicolumn{8}{l}{\cellcolor{lightgray}0-Shot-CoT} \\
            Llama2-70b-chat & 3.5 & 69.7 & 64.3 & 77.1 & 80.0 & 11.3  & 13.9\\
            GPT-4 & 10.4 & 75.1 & 73.0 & 74.3 & 82.9 & 25.2 & 34.8\\
            \multicolumn{8}{l}{\cellcolor{lightgray}\textsc{TimeToM}\xspace} \\
            \multirow{3}{*}{Llama2-70b-chat} & 6.1 & 79.0 & 75.7 & 80.0 & 88.6  & 17.4 &  15.7 \\ 
            & {\textcolor[RGB]{0,201,87} {\textbf{(+6.1, +2.6)}}} & \textbf{(+72.5, +9.3)} & \textbf{(+67.0, +11.4)} & \textbf{(+80.0, +2.9)} & \textbf{(+82.9, +8.6)} & {\textcolor[RGB]{0,201,87} {\textbf{(+13.1, +6.1)}}} &  {\textcolor[RGB]{0,201,87} {\textbf{(+7.0, +1.8)}}}\\
            & {\textcolor[RGB]{0,201,87} {\textbf{($\times$ $\infty$, $\times$ 1.7)}}} & & & & & {\textcolor[RGB]{0,201,87} {\textbf{($\times$ 4.0, $\times$ 1.5)}}}& {\textcolor[RGB]{0,201,87} {\textbf{($\times$ 1.8, $\times$ 1.1)}}}\\
            \multirow{3}{*}{GPT-4} & 41.7 & 93.0 & 93.1 & 94.3 & 91.5 & 51.3  & 52.2\\
            & {\textcolor[RGB]{0,201,87} {\textbf{(+33.0, +31.3)}}} & \textbf{(+16.8, +17.9)} & \textbf{(+20.1, +20.1)} & \textbf{(+17.2, +20.0)} & \textbf{(+5.8, +8.6)} & {\textcolor[RGB]{0,201,87} {\textbf{(+27.8, +26.1)}}} & {\textcolor[RGB]{0,201,87} {\textbf{(+28.7, +17.3)}}}\\
            & {\textcolor[RGB]{0,201,87} {\textbf{($\times$ 4.8, $\times$ 4.0)}}} & & & & & {\textcolor[RGB]{0,201,87} {\textbf{($\times$ 2.2, $\times$ 2.0)}}}&{\textcolor[RGB]{0,201,87} {\textbf{($\times$ 2.2, $\times$ 1.5)}}}\\
            \midrule
            Avg & {\textcolor[RGB]{0,201,87} {\textbf{(+19.6, +17.0)}}} & \textbf{(+44.7, +13.6)} & \textbf{(+43.6, +15.8)} & \textbf{(+48.6, +11.5)} & \textbf{(+44.4, +8.6)} & {\textcolor[RGB]{0,201,87} {\textbf{(+20.5, +16.1)}}} & {\textcolor[RGB]{0,201,87} {\textbf{(+17.9, +9.6)}}}\\



            \bottomrule
    \end{tabular}
    \end{adjustbox}
    \caption{\textsc{TimeToM}\xspace results on FanToM. We present \textbf{absolute accuracy difference} between \textsc{TimeToM}\xspace and the baselines (0-shot and 0-shot-CoT), and {\textcolor[RGB]{0,201,87} {\textbf{green}}} will appear on metrics related to the {\textcolor[RGB]{0,201,87} {\textbf{robustness}}} of ToM reasoning. Results for list-type and binary-type questions can be found in Appendix \ref{appendix3}.
    }
    \vspace{-10pt}
    \label{tab:2}
\end{center}\end{table*}}
\paragraph{Better ToM Reasoning Robustness} 
We use \textsc{All*} and All score from Table \ref{tab:2} to evaluate the ToM reasoning robustness of baselines and \textsc{TimeToM}\xspace. We achieve {\textcolor[RGB]{0,201,87} {\textbf{+33.0\% ($\times$ 4.8)}}} and {\textcolor[RGB]{0,201,87} {\textbf{+31.3\% ($\times$ 4.0)}}} absolute accuracy improvement over the 0-shot and 0-shot-CoT GPT-4 baselines for \textsc{All*} score, which requires correct answers to all five types of ToM questions (Belief, Answerability[List, Y/N], and Infoaccess[List, Y/N]). For All scores under the Answerability question and Infoaccess question, which require correct answers to both list-type and Y/N-type questions, we achieve {\textcolor[RGB]{0,201,87} {\textbf{+27.8\% ($\times$ 2.2)}}} and {\textcolor[RGB]{0,201,87} {\textbf{+26.1\% ($\times$ 2.0)}}}, as well as {\textcolor[RGB]{0,201,87} {\textbf{+28.7\% ($\times$ 2.2)}}} and {\textcolor[RGB]{0,201,87} {\textbf{+17.3\% ($\times$ 1.5)}}} absolute accuracy improvement over the 0-shot and 0-shot-CoT GPT-4 baselines, respectively. A similar improvement is noticeable in the llama2-70b-chat model, although the degree of improvement is not as large. 


\section{The Effect of Key Components}
Given \textsc{TimeToM}'s strong performance, we analyze its key components: (1) Foundation: constructing temporal space. (2) From the perspective of first-order and higher-order ToM questions, considering temporal belief construction and compression as well as time-aware belief solver.
\paragraph{Constructing Temporal Space}
The construction of temporal space provides LLMs with a clearer understanding of object states and character actions, especially for those models with weaker cognitive abilities. We conduct experiments on the Llama2 series models in 0-shot and 0-shot with timeline settings, results show that the construction of the temporal space has led to significant performance improvements in true-belief and mem-real questions, which are associated with the real-world state. Moreover, the clear cognition brought by the temporal space is also helpful for reasoning about false belief questions. Detailed results data can be found in Appendix \ref{appendix_effect}. Case 1 in Figure \ref{fig:example2} vividly illustrates the benefits of constructing temporal space.

\paragraph{Temporal Belief Construction and Compression} Within the temporal space, we construct TBSC for the characters and utilize self-world belief, obtained through belief compression, to answer first-order ToM questions. By comparing the performance of the Llama2-7b-chat and Llama2-13b-chat models on first-order ToM questions in Tables \ref{tab:1} and \ref{tab:accuracy}, an observed improvement of \textbf{+3.75\%} and \textbf{+2.75\%} brought by belief construction and compression is noted. A larger improvement \textbf{+10.50\%} and \textbf{+29.50\%} appears in the GPT-3.5-turbo and GPT-4 models, as they inherently possess stronger language comprehension abilities, construct better TBSCs, perform more effective belief compression.

\begin{figure*}
    \centering
    \includegraphics[width=0.93\textwidth]{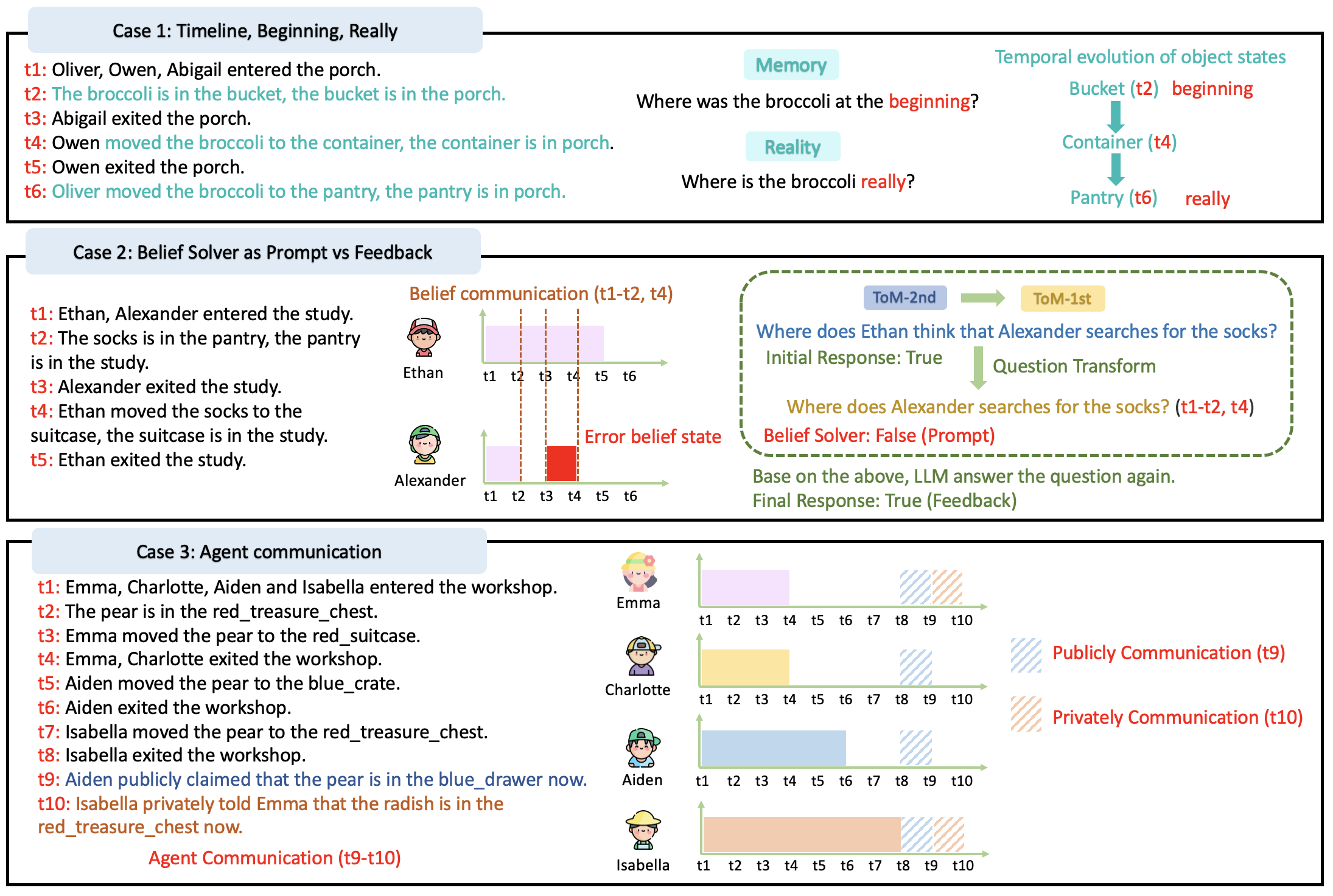}
    \caption{Case 1: The benefit of constructing temporal space. Case 2: The comparison between the belief solver as prompt and as feedback. Case 3: The application of \textsc{TimeToM}\xspace in situations involving agent communication.}
    \label{fig:example2}
    \vspace{-3mm}
\end{figure*}

\paragraph{Tool—Time-Aware Belief Solver}
Within the temporal space, we design a novel tool—belief solver to inspire the reasoning process of LLMs on higher-order ToM questions. As shown in Table \ref{tab:accuracy4}, the incorporation of belief solver results in a performance improvement of \textbf{+7.0\%} and \textbf{+16.0\%} for the GPT-3.5-turbo and GPT-4 models, respectively.

\definecolor{de}{HTML}{f4cbbd}
\definecolor{in}{HTML}{c8daea}

\begin{table}[H]
    \centering
    \small
    \begin{tabular}{cccc}
    \toprule
    Model & Response & Result & Relation\\
    \midrule
    \multirow{3}{*}{GPT-3.5-turbo} & Initial & 65.0 & \multirow{1.2}{*}{\scriptsize \colorbox{de}{Collaboration}}\\ & Tool & 69.0 & \multirow{2}{*}{\scriptsize \colorbox{de}{+7.0}}\\ & Final & 72.0 & \\
    \midrule
    \multirow{3}{*}{GPT-4} & Initial & 79.0 & \multirow{1.2}{*}{\scriptsize \colorbox{in}{Tool Dominates}}\\ & Tool & 96.0 & \multirow{2}{*}{\scriptsize \colorbox{in}{+16.0}}\\ & Final & 95.0 & \\
    
    
    
    
    
    
    \bottomrule
    \end{tabular}
    \caption{Comparison of initial, tool (belief solver as prompt), and final (belief solver as feedback) response performance of GPT series models on higher-order ToM questions under ToMI benchmark.}
    \label{tab:accuracy4}
\end{table}


\section{Analysis}
In this section, we analyze belief solver as prompt vs feedback and the extension of \textsc{TimeToM}\xspace to situations encompassing agent communication.

\paragraph{Belief Solver as Prompt vs Feedback}
Given that the belief solver can transform higher-order ToM questions into first-order ToM questions under belief communication periods, why don't we use it directly as a prompt when answering higher-order ToM questions? There are two reasons: (1) Given the GPT-4 model's exceptional language comprehension capabilities, it can construct accurate TBSC for each character. Consequently, it can accurately determine periods of belief communication through intersection operations of TBSC between characters. But as the model's ability decreases, e.g., for the GPT-3.5-turbo model, the TBSC of the characters it builds will have a certain probability of error, and then the probability of obtaining an incorrect belief communication periods is greatly increased when performing the TBSC intersection operation between characters. (2) It is more effective to use the belief solver as feedback to inspire the reasoning process of LLMs on higher-order ToM questions. LLMs will consider the initial reasoning perspective and the reasoning perspective provided by the feedback to form a final response, and when the reasoning perspective provided by the feedback is accurate and effective, the LLMs also acknowledge this perspective, which corresponds to the GPT-4 model case. Conversely, if the reasoning perspective offered by the feedback is erroneous, the LLMs have a certain probability of recognizing this error and realizing the integration of useful information from both perspectives to achieve better performance, which corresponds to the GPT-3.5-turbo model case. Case 2 in Figure \ref{fig:example2} and Table \ref{tab:accuracy4} offers qualitative and quantitative analysis support for the above two reasons.

\paragraph{Applicable to Situations Involving Communication between Agents.}

In real-world interactions, people engage in sharing their innermost thoughts with each other, including both their perceptions of situations and observations of others' behaviors. \citet{he2023hi} recently propose a benchmark encompassing agent communication, considering \textsc{TimeToM}\xspace in this situation, which can model agent communication as a belief communication between agents at a specific time point. As shown in case 3 of Fig \ref{fig:example2}, \textsc{TimeToM}\xspace has good applicability to situations involving agent communication. 




\section{Conclusion}
In this paper, we propose \textsc{TimeToM}\xspace to improve the ToM capabilities of LLMs in reading comprehension and interactive dialogue scenarios. Specifically, we first construct temporal space which serves as the foundation. Building on this, we develop several key components: character's belief state chain construction, social world model cognition-inspired belief compression, and tool—belief solver. Extensive experimental results show that \textsc{TimeToM}\xspace substantially improves the reasoning performance of LLMs on ToM questions while making a significant advance towards coherent and robust ToM reasoning. The temporal space, serving as a key, unlocks the door to the LLMs' Theory of Mind. Furthermore, we find that \textsc{TimeToM}\xspace can also be extended to situations involving agent communication which is commonly occurring in the real-world. 

\section*{Limitations}
There are two major limitations in \textsc{TimeToM}\xspace. Firstly, the belief solver relies on constructing an accurate TBSC for characters. We conduct experiments on models with a parameter scale of 7B or larger. For models with less than 7B parameters, due to their relatively weaker instruction understanding ability, the error rate in constructing TBSCs is higher, which in turn affects the effectiveness of the belief solver. However, with the continuous development of LLMs, this limitation can be solved easily. Secondly, we focus on ToM reasoning for textual modality, it is also important to perform effective multimodal ToM reasoning, which we treat as future work.

\section*{Acknowledgements}
This work is supported by the National Natural Science Foundation of China (No. 62376245), the Key Research and Development Program of Zhejiang Province, China (No. 2024C01034),  the National Natural Science Foundation of Zhejiang Province (LY24C090001) and the Fundamental Research Funds for the Central Universities.

\bibliography{anthology,custom}

\begin{thebibliography}{32}
\expandafter\ifx\csname natexlab\endcsname\relax\def\natexlab#1{#1}\fi

\bibitem[{Achiam et~al.(2023)Achiam, Adler, Agarwal, Ahmad, Akkaya, Aleman, Almeida, Altenschmidt, Altman, Anadkat et~al.}]{achiam2023gpt}
Josh Achiam, Steven Adler, Sandhini Agarwal, Lama Ahmad, Ilge Akkaya, Florencia~Leoni Aleman, Diogo Almeida, Janko Altenschmidt, Sam Altman, Shyamal Anadkat, et~al. 2023.
\newblock Gpt-4 technical report.
\newblock \emph{arXiv preprint arXiv:2303.08774}.

\bibitem[{Arodi and Cheung(2021)}]{arodi2021textual}
Akshatha Arodi and Jackie Chi~Kit Cheung. 2021.
\newblock Textual time travel: A temporally informed approach to theory of mind.
\newblock In \emph{Findings of the Association for Computational Linguistics: EMNLP 2021}, pages 4162--4172.

\bibitem[{Frith and Frith(2003)}]{frith2003development}
Uta Frith and Christopher~D Frith. 2003.
\newblock Development and neurophysiology of mentalizing.
\newblock \emph{Philosophical Transactions of the Royal Society of London. Series B: Biological Sciences}, 358(1431):459--473.

\bibitem[{Gandhi et~al.(2023)Gandhi, Fr{\"a}nken, Gerstenberg, and Goodman}]{gandhi2023understanding}
Kanishk Gandhi, Jan-Philipp Fr{\"a}nken, Tobias Gerstenberg, and Noah~D Goodman. 2023.
\newblock Understanding social reasoning in language models with language models.
\newblock \emph{arXiv preprint arXiv:2306.15448}.

\bibitem[{Gandhi et~al.(2021)Gandhi, Stojnic, Lake, and Dillon}]{gandhi2021baby}
Kanishk Gandhi, Gala Stojnic, Brenden~M Lake, and Moira~R Dillon. 2021.
\newblock Baby intuitions benchmark (bib): Discerning the goals, preferences, and actions of others.
\newblock \emph{Advances in neural information processing systems}, 34:9963--9976.

\bibitem[{Hao et~al.(2023)Hao, Gu, Ma, Hong, Wang, Wang, and Hu}]{hao2023reasoning}
Shibo Hao, Yi~Gu, Haodi Ma, Joshua~Jiahua Hong, Zhen Wang, Daisy~Zhe Wang, and Zhiting Hu. 2023.
\newblock Reasoning with language model is planning with world model.
\newblock \emph{arXiv preprint arXiv:2305.14992}.

\bibitem[{He et~al.(2023)He, Wu, Jia, Mihalcea, Chen, and Deng}]{he2023hi}
Yinghui He, Yufan Wu, Yilin Jia, Rada Mihalcea, Yulong Chen, and Naihao Deng. 2023.
\newblock Hi-tom: A benchmark for evaluating higher-order theory of mind reasoning in large language models.
\newblock \emph{arXiv preprint arXiv:2310.16755}.

\bibitem[{He-Yueya et~al.(2023)He-Yueya, Poesia, Wang, and Goodman}]{he2023solving}
Joy He-Yueya, Gabriel Poesia, Rose~E Wang, and Noah~D Goodman. 2023.
\newblock Solving math word problems by combining language models with symbolic solvers.
\newblock \emph{arXiv preprint arXiv:2304.09102}.

\bibitem[{Hu and Shu(2023)}]{hu2023language}
Zhiting Hu and Tianmin Shu. 2023.
\newblock Language models, agent models, and world models: The law for machine reasoning and planning.
\newblock \emph{arXiv preprint arXiv:2312.05230}.

\bibitem[{Kim et~al.(2023)Kim, Sclar, Zhou, Bras, Kim, Choi, and Sap}]{kim2023fantom}
Hyunwoo Kim, Melanie Sclar, Xuhui Zhou, Ronan~Le Bras, Gunhee Kim, Yejin Choi, and Maarten Sap. 2023.
\newblock Fantom: A benchmark for stress-testing machine theory of mind in interactions.
\newblock \emph{arXiv preprint arXiv:2310.15421}.

\bibitem[{Le et~al.(2019)Le, Boureau, and Nickel}]{le2019revisiting}
Matthew Le, Y-Lan Boureau, and Maximilian Nickel. 2019.
\newblock Revisiting the evaluation of theory of mind through question answering.
\newblock In \emph{Proceedings of the 2019 Conference on Empirical Methods in Natural Language Processing and the 9th International Joint Conference on Natural Language Processing (EMNLP-IJCNLP)}, pages 5872--5877.

\bibitem[{Ma et~al.(2023)Ma, Sansom, Peng, and Chai}]{ma2023towards}
Ziqiao Ma, Jacob Sansom, Run Peng, and Joyce Chai. 2023.
\newblock Towards a holistic landscape of situated theory of mind in large language models.
\newblock \emph{arXiv preprint arXiv:2310.19619}.

\bibitem[{Madaan et~al.(2023)Madaan, Tandon, Gupta, Hallinan, Gao, Wiegreffe, Alon, Dziri, Prabhumoye, Yang et~al.}]{madaan2023self}
Aman Madaan, Niket Tandon, Prakhar Gupta, Skyler Hallinan, Luyu Gao, Sarah Wiegreffe, Uri Alon, Nouha Dziri, Shrimai Prabhumoye, Yiming Yang, et~al. 2023.
\newblock Self-refine: Iterative refinement with self-feedback.
\newblock \emph{arXiv preprint arXiv:2303.17651}.

\bibitem[{Mitchell(2023)}]{doi:10.1126/science.adm8175}
Melanie Mitchell. 2023.
\newblock \href {https://doi.org/10.1126/science.adm8175} {Ai’s challenge of understanding the world}.
\newblock \emph{Science}, 382(6671):eadm8175.

\bibitem[{Nathani et~al.(2023)Nathani, Wang, Pan, and Wang}]{nathani2023maf}
Deepak Nathani, David Wang, Liangming Pan, and William~Yang Wang. 2023.
\newblock Maf: Multi-aspect feedback for improving reasoning in large language models.
\newblock \emph{arXiv preprint arXiv:2310.12426}.

\bibitem[{Nematzadeh et~al.(2018)Nematzadeh, Burns, Grant, Gopnik, and Griffiths}]{nematzadeh2018evaluating}
Aida Nematzadeh, Kaylee Burns, Erin Grant, Alison Gopnik, and Thomas~L Griffiths. 2018.
\newblock Evaluating theory of mind in question answering.
\newblock \emph{arXiv preprint arXiv:1808.09352}.

\bibitem[{Premack and Woodruff(1978)}]{premack1978does}
David Premack and Guy Woodruff. 1978.
\newblock Does the chimpanzee have a theory of mind?
\newblock \emph{Behavioral and brain sciences}, 1(4):515--526.

\bibitem[{Quesque and Rossetti(2020)}]{quesque2020theory}
Fran{\c{c}}ois Quesque and Yves Rossetti. 2020.
\newblock What do theory-of-mind tasks actually measure? theory and practice.
\newblock \emph{Perspectives on Psychological Science}, 15(2):384--396.

\bibitem[{Rabinowitz et~al.(2018)Rabinowitz, Perbet, Song, Zhang, Eslami, and Botvinick}]{rabinowitz2018machine}
Neil Rabinowitz, Frank Perbet, Francis Song, Chiyuan Zhang, SM~Ali Eslami, and Matthew Botvinick. 2018.
\newblock Machine theory of mind.
\newblock In \emph{International conference on machine learning}, pages 4218--4227. PMLR.

\bibitem[{Sap et~al.(2022)Sap, LeBras, Fried, and Choi}]{sapNeuralTheoryofMindLimits2022}
Maarten Sap, Ronan LeBras, Daniel Fried, and Yejin Choi. 2022.
\newblock Neural theory-of-mind? on the limits of social intelligence in large lms.
\newblock \emph{arXiv preprint arXiv:2210.13312}.

\bibitem[{Sclar et~al.(2023)Sclar, Kumar, West, Suhr, Choi, and Tsvetkov}]{sclar2023minding}
Melanie Sclar, Sachin Kumar, Peter West, Alane Suhr, Yejin Choi, and Yulia Tsvetkov. 2023.
\newblock Minding language models'(lack of) theory of mind: A plug-and-play multi-character belief tracker.
\newblock \emph{arXiv preprint arXiv:2306.00924}.

\bibitem[{Sclar et~al.(2022)Sclar, Neubig, and Bisk}]{sclar2022symmetric}
Melanie Sclar, Graham Neubig, and Yonatan Bisk. 2022.
\newblock Symmetric machine theory of mind.
\newblock In \emph{International Conference on Machine Learning}, pages 19450--19466. PMLR.

\bibitem[{Shapira et~al.(2023)Shapira, Levy, Alavi, Zhou, Choi, Goldberg, Sap, and Shwartz}]{shapira2023clever}
Natalie Shapira, Mosh Levy, Seyed~Hossein Alavi, Xuhui Zhou, Yejin Choi, Yoav Goldberg, Maarten Sap, and Vered Shwartz. 2023.
\newblock Clever hans or neural theory of mind? stress testing social reasoning in large language models.
\newblock \emph{arXiv preprint arXiv:2305.14763}.

\bibitem[{Shinn et~al.(2023)Shinn, Cassano, Gopinath, Narasimhan, and Yao}]{shinn2023reflexion}
Noah Shinn, Federico Cassano, Ashwin Gopinath, Karthik~R Narasimhan, and Shunyu Yao. 2023.
\newblock Reflexion: Language agents with verbal reinforcement learning.
\newblock In \emph{Thirty-seventh Conference on Neural Information Processing Systems}.

\bibitem[{Shu et~al.(2021)Shu, Bhandwaldar, Gan, Smith, Liu, Gutfreund, Spelke, Tenenbaum, and Ullman}]{shu2021agent}
Tianmin Shu, Abhishek Bhandwaldar, Chuang Gan, Kevin Smith, Shari Liu, Dan Gutfreund, Elizabeth Spelke, Joshua Tenenbaum, and Tomer Ullman. 2021.
\newblock Agent: A benchmark for core psychological reasoning.
\newblock In \emph{International Conference on Machine Learning}, pages 9614--9625. PMLR.

\bibitem[{Touvron et~al.(2023)Touvron, Martin, Stone, Albert, Almahairi, Babaei, Bashlykov, Batra, Bhargava, Bhosale et~al.}]{touvron2023llama}
Hugo Touvron, Louis Martin, Kevin Stone, Peter Albert, Amjad Almahairi, Yasmine Babaei, Nikolay Bashlykov, Soumya Batra, Prajjwal Bhargava, Shruti Bhosale, et~al. 2023.
\newblock Llama 2: Open foundation and fine-tuned chat models.
\newblock \emph{arXiv preprint arXiv:2307.09288}.

\bibitem[{Ullman(2023)}]{ullman2023large}
Tomer Ullman. 2023.
\newblock Large language models fail on trivial alterations to theory-of-mind tasks.
\newblock \emph{arXiv preprint arXiv:2302.08399}.

\bibitem[{Wei et~al.(2022)Wei, Wang, Schuurmans, Bosma, Xia, Chi, Le, Zhou et~al.}]{wei2022chain}
Jason Wei, Xuezhi Wang, Dale Schuurmans, Maarten Bosma, Fei Xia, Ed~Chi, Quoc~V Le, Denny Zhou, et~al. 2022.
\newblock Chain-of-thought prompting elicits reasoning in large language models.
\newblock \emph{Advances in Neural Information Processing Systems}, 35:24824--24837.

\bibitem[{Wilf et~al.(2023)Wilf, Lee, Liang, and Morency}]{wilf2023think}
Alex Wilf, Sihyun~Shawn Lee, Paul~Pu Liang, and Louis-Philippe Morency. 2023.
\newblock Think twice: Perspective-taking improves large language models' theory-of-mind capabilities.
\newblock \emph{arXiv preprint arXiv:2311.10227}.

\bibitem[{Yao et~al.(2023)Yao, Yu, Zhao, Shafran, Griffiths, Cao, and Narasimhan}]{yao2023tree}
Shunyu Yao, Dian Yu, Jeffrey Zhao, Izhak Shafran, Thomas~L Griffiths, Yuan Cao, and Karthik Narasimhan. 2023.
\newblock Tree of thoughts: Deliberate problem solving with large language models.
\newblock \emph{arXiv preprint arXiv:2305.10601}.

\bibitem[{Yue(2022)}]{yue2022world}
Yutao Yue. 2022.
\newblock A world-self model towards understanding intelligence.
\newblock \emph{IEEE Access}, 10:63034--63048.

\bibitem[{Zhou et~al.(2022)Zhou, Sch{\"a}rli, Hou, Wei, Scales, Wang, Schuurmans, Cui, Bousquet, Le et~al.}]{zhou2022least}
Denny Zhou, Nathanael Sch{\"a}rli, Le~Hou, Jason Wei, Nathan Scales, Xuezhi Wang, Dale Schuurmans, Claire Cui, Olivier Bousquet, Quoc Le, et~al. 2022.
\newblock Least-to-most prompting enables complex reasoning in large language models.
\newblock \emph{arXiv preprint arXiv:2205.10625}.

\end{thebibliography}
\bibliographystyle{acl_natbib}

\clearpage
\appendix

\noindent\textbf{Appendix}

\section{Benchmark Details and Evaluation Metrics}
\label{sec:appendix1}
\subsection{ToMI}
ToMI \cite{le2019revisiting} is a benchmark in the reading comprehension scenarios, strictly imitating the Sally-Anne test, including the story, questions, and answer choices. The structure of the story is as follows: 

\begin{figure}[H]
    \centering
    \includegraphics[width=0.36\textwidth]{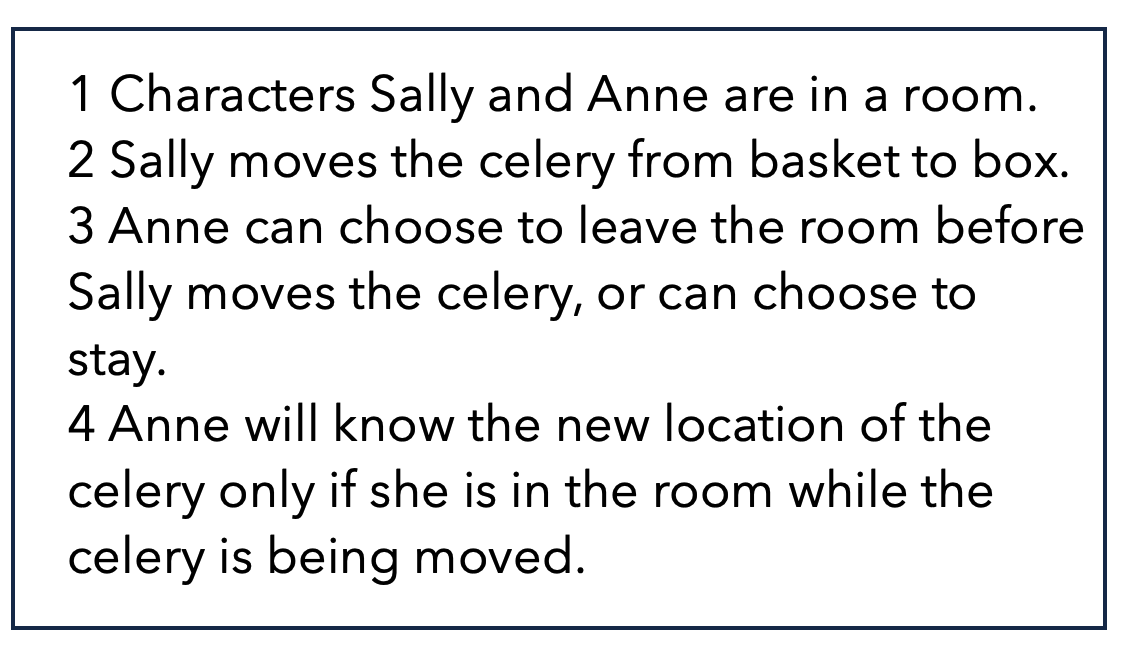}
    \caption{Story structure of ToMI.}
    \label{fig:app1}
    \vspace{-3mm}
\end{figure}

\noindent Five types of ToM questions are proposed: first-order or second-order, exploring characters' true or false beliefs (i.e., beliefs that are consistent or inconsistent with reality) as well as questions exploring reality and memory (zeroth-order ToM, \cite{sclar2022symmetric}). The formatted description for each type of question is as follows:

\begin{description}
    \item[Reality:] Where is celery really?
    \item[Memory:] Where was celery at the beginning?
    \item[First-Order Belief Sally:] Where will Sally look for celery?
    \item[First-Order Belief Anne:] Where will Anne look for celery?
    \item[Second-Order Belief Sally:] Where does Sally believe Anne will look for celery?
    \item[Second-Order Belief Anne:] Where does Anne believe Sally will look for celery?
\end{description}

\noindent Meanwhile, in \textbf{first-order belief} and \textbf{second-order belief} questions, both \textbf{false belief} and \textbf{true belief} are involved. For example: Sally moves the celery from basket to box without Anne observing this action. A \textbf{first-order belief} question: "Where will Anne look for celery?" Since Sally has moved the celery, Anne’s belief will be incorrect – this type of question is called \textbf{false belief} and has its counterpart in \textbf{true belief} questions, where Anne’s belief about the world is correct. 

An updated version of ToMI proposed from (\citet{arodi2021textual, sapNeuralTheoryofMindLimits2022}) that has relabelled mislabelled second-order questions and disambiguated the location of containers after their reference. \citet{sclar2023minding} expands the story structure by introducing more characters and containers. 

All questions have two possible answers: the \textbf{original O location}, and the \textbf{final O location}. ToMI is a binary multiple-choice task, with random accuracy being 50\%. In our experiments, we select the exact same test dataset used in \textsc{SimToM}\xspace (i.e., the updated version of ToMI) to ensure a fair comparison.

\subsection{BigToM}
BigToM \cite{gandhi2023understanding} is another benchmark in the reading comprehension scenario and also follows the Sally-Anne test format. It is generated by GPT-4. Unlike ToMI, BigToM tells stories in more natural language and is not limited to changes in object locations. In our experiments, we focus on the "Forward belief" questions rather than "Backward belief" to align more closely with the structure of ToMI questions. The definitions of \textbf{true belief} and \textbf{false belief} questions in BigToM are the same as in ToMI. BigToM is also a binary multiple-choice task, with random accuracy being 50\%.

\subsection{FanToM}
\begin{figure*}
    \centering
    \includegraphics[width=0.85\textwidth]{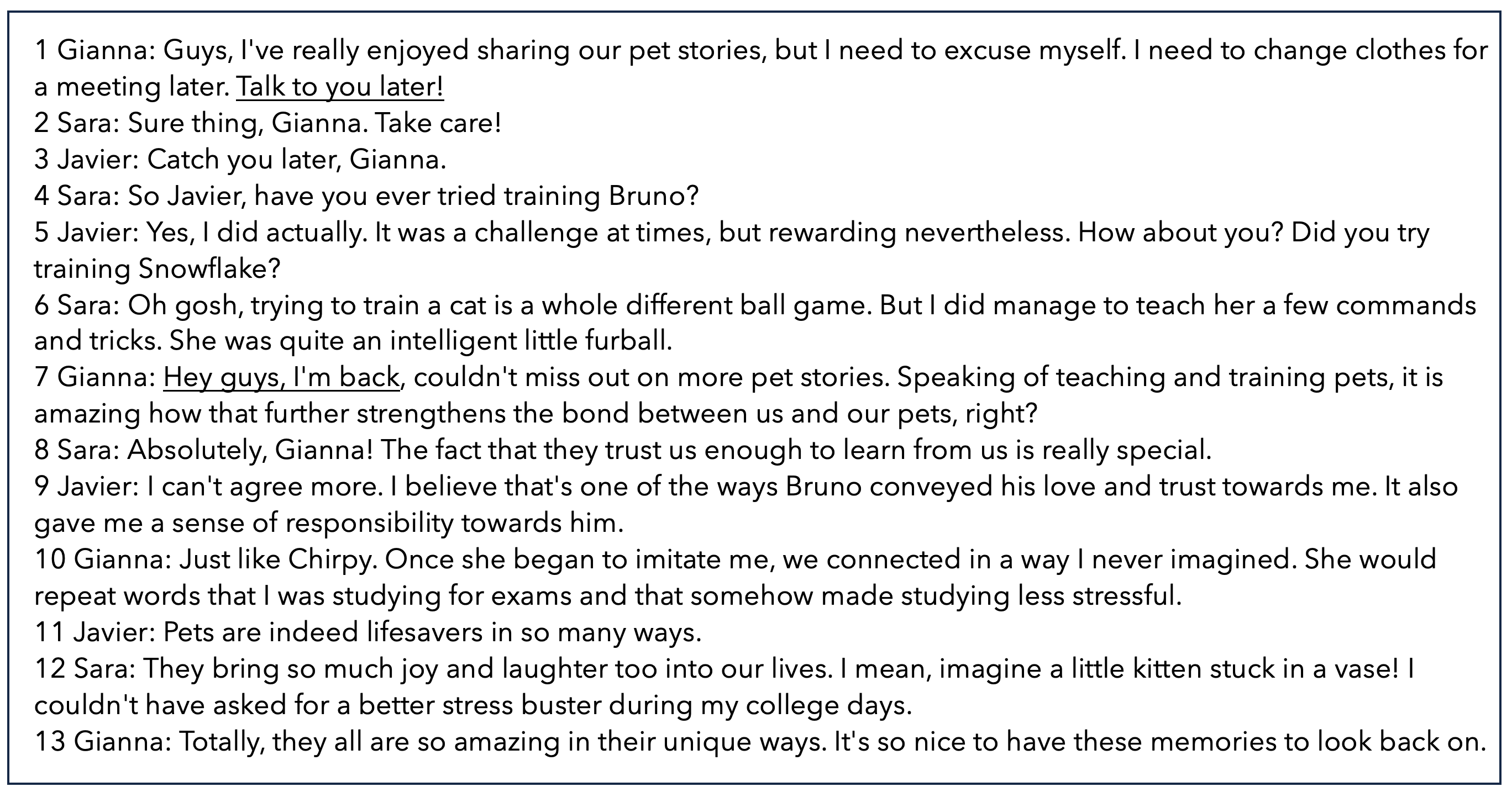}
    \caption{Dialogue structure of FanToM.}
    \label{fig:app2}
    \vspace{-3mm}
\end{figure*}

FanToM \cite{kim2023fantom} is a benchmark in interactive dialogue scenarios. Dialogues involve information asymmetry \cite{quesque2020theory}, with characters joining and leaving the dialogues while it continues, to simulate distinct mental states. The structure of dialogue is illustrated in Figure \ref{fig:app2}.
The authors of FanToM formulate multiple types of questions (belief questions, answerability questions[list, binary], and infoaccess questions[list, binary]) that demand the same underlying reasoning to identify the illusory or false sense of ToM capabilities in LLMs:
\begin{description}
    \item[First-Order Belief:] What does Gianna believe about who discussed their experiences training their pets, Bruno and Snowflake?
    \item[Third-Order Acyclic Belief:] What does Sara believe about Javier's belief regarding Gianna believe about who discussed their experiences training their pets, Bruno and Snowflake?
    \item[Third-Order Cyclic Belief:] What does Sara believe about Javier's belief regarding Sara believe about who discussed their experiences training their pets, Bruno and Snowflake?
    \item[Answerability\_List:] Who discussed their experiences training their pets, Bruno and Snowflake? List all the characters who know the precise correct answer to this question.
    \item[Answerability\_Binary] Who discussed their experiences training their pets, Bruno and Snowflake? Does Gianna (Javier, Sara) know the precise correct answer to this question?
    \item[Infoaccess\_List:] Sara and Javier discussed their experiences training their pets, Bruno and Snowflake. List all the characters who know this information.
    \item[Infoaccess\_Binary:] Sara and Javier discussed their experiences training their pets, Bruno and Snowflake. Does Gianna (Javier, Sara) know this information?
\end{description}
In our experiments, we focus not only on the performance of our method in answering individual question types but also on its ability to reason coherently and robustly across multiple question types. All questions have two possible answers, with random accuracy being 50\%.

\subsection{Evaluation Metrics}
Following \citet{wilf2023think}, we report accuracy for all questions under ToMI and BigToM. Following \citet{kim2023fantom}, for FanToM, we report accuracy for belief, answerability[list] and infoaccess[list] questions. The weighted F1 scores are reported for answerability[binary] and infoaccess[binary] questions. To evaluate the reasoning robustness of LLMs on ToM questions, we report the All score for answerability and infoaccess questions requiring models to be correct on both list-type and binary-type questions, the \textsc{All*} score which requires the models to answer all five ToM question types which require the same type of ToM reasoning.



\section{Our Prompt}\label{Our prompt}
\subsection{Prompt for Interactive Dialogue Scenario}
\label{appendix1}
\lstset{
  breaklines=true, 
  basicstyle=\ttfamily\small, 
  breakindent=0pt,
  escapeinside={(*@}{@*)}, 
}


\begin{lstlisting}
(*@\textbf{Constructing Temporal Space}@*): 
The following is a dialogue. Your task is to add timeline to the dialogue.

Here are one rules: Each utterance spoken by a character corresponds to a moment t, Use \n as a delimiter, and the timeline is t1,t2,... ,tN.

Dialogue:
{(*@\color{blue}{dialogue}@*)}

Only output the dialogue content with the added timeline, do not provide explanations.
\end{lstlisting}

\begin{lstlisting}
(*@\textbf{Temporal Belief State Chain Construction}@*): 
The following is a dialogue with a timeline between multiple characters. 
Your task is to only output the dialogue content on the timeline that the character {(*@\color{blue}{character}@*)} can aware of.

Here are two rules:
If a character leaves the conversation to do something else and then back after a few rounds of dialogue, they are unaware of the content of the conversation that took place during their absence, but they aware of the content of the conversation besides their absence.
If a character don't leaves the conversation to do something else and then back after a few rounds of dialogue. They are aware of all the content of dialogue with all timeline.

Dialogue:
{(*@\color{blue}{dialogue}@*)}

What dialogue content on the timeline does {(*@\color{blue}{character}@*)} aware of? Only output the dialogue content according to the above rules, do not provide an explanation.
\end{lstlisting}

\begin{lstlisting}
(*@\textbf{Time-Aware Belief Question Answer with Belief Compression (First-order ToM questions})@*): 
The following is the belief states chain of character {(*@\color{blue}{name}@*)}. This is the content known to {(*@\color{blue}{name}@*)}:[{(*@\color{blue}{perspective}@*)}]
You are {(*@\color{blue}{name}@*)}.
Based on the above information, answer the following question:
{(*@\color{blue}{question}@*)}
When answering questions, based on own belief, simply focus on the information of things asked in the question and ignore other distracting factors. You must choose one of the above choices.
\end{lstlisting}

\begin{lstlisting}
(*@\textbf{Time-Aware Belief Question Answer without Belief Compression (Higher-order ToM questions)}@*): 
The following is the belief states chain of character {(*@\color{blue}{name}@*)}. This is the content known to {(*@\color{blue}{name}@*)}:[{(*@\color{blue}{perspective}@*)}]
You are {(*@\color{blue}{name}@*)}.
Based on the above information, answer the following question:
{(*@\color{blue}{question}@*)}
You must choose one of the above choices.
\end{lstlisting}

\begin{lstlisting}
(*@\textbf{Time-Aware Answerablity Question[List] Answer}@*): 
The following is the belief states chain of each character. This is the content known to each character.
Each character only knows the contents within their own belief state chain and is unaware of the contents within the belief state chain of other characters.
{(*@\color{blue}{final\_text}@*)}
Question:
{(*@\color{blue}{target}@*)}
Based on the belief state chain of the above-mentioned characters, only output all the characters who know the precise correct answer to this question, do not provide an explanation.
\end{lstlisting}

\begin{lstlisting}
(*@\textbf{Time-Aware Answerablity Question[Binary] Answer}@*): 
The following is the belief states chain of character {(*@\color{blue}{character}@*)}. This is the content known to {character}.
{(*@\color{blue}{binary\_context}@*)}
Question:
{(*@\color{blue}{target}@*)}
Based on the belief state chain of character {(*@\color{blue}{character}@*)}, does {(*@\color{blue}{character}@*)} know the precise correct answer to this question? Answer yes or no. Answer:.
\end{lstlisting}

\begin{lstlisting}
(*@\textbf{Time-Aware Infoaccess Question[List] Answer}@*): 
The following is the belief states chain of each character. This is the content known to each character.
Each character only knows the contents within their own belief state chain and is unaware of the contents within the belief state chain of other characters.
{(*@\color{blue}{final\_text}@*)}
Target:
{(*@\color{blue}{target\_q}@*)}
{(*@\color{blue}{target\_a}@*)}
Question:
Based on the belief state chain of the above-mentioned characters, only output all the characters who know the target information, do not provide an explanation.
\end{lstlisting}

\begin{lstlisting}
(*@\textbf{Time-Aware Infoaccess Question[Binary] Answer}@*): 
The following is the belief states chain of character {(*@\color{blue}{character}@*)}. This is the content known to {(*@\color{blue}{character}@*)}.
{(*@\color{blue}{binary\_context}@*)}
Target:
{(*@\color{blue}{target\_q}@*)}
{(*@\color{blue}{target\_a}@*)}
Question:
Based on the belief state chain of character {(*@\color{blue}{character}@*)}, does {(*@\color{blue}{character}@*)} know the target information? Answer yes or no. Answer:.
\end{lstlisting}

\begin{lstlisting}
(*@\textbf{Time-Aware Belief Solver}@*): 
The following is the belief states chain of character {(*@\color{blue}{name}@*)}. This is the content known to {(*@\color{blue}{name}@*)}:[{(*@\color{blue}{perspective}@*)}]
You are {(*@\color{blue}{name}@*)}.
Based on the above information, answer the following question:
{(*@\color{blue}{question}@*)}
Answer:{(*@\color{blue}{answer}@*)}
Feedback: The event corresponding to the period of belief communication between characters {(*@\color{blue}{character1}@*)}, {(*@\color{blue}{character2}@*)} and {(*@\color{blue}{character3}@*)}: {(*@\color{blue}{common\_belief}@*)} Based on this information, the answer we get to the question:{(*@\color{blue}{question}@*)} is [{(*@\color{blue}{answer2}@*)}]
Considering this feedback, answer the question: {(*@\color{blue}{question}@*)} again. Keep your answer concise, one sentence is enough. You must choose one of the above choices.
\end{lstlisting}

\subsection{Prompt for Reading Comprehension Scenario}
\label{appendix2}
\begin{lstlisting}
(*@\textbf{Constructing Temporal Space}@*): 
The following is a story. Your task is to add timeline to the story.

Here are one rules: Each sentence corresponds to a moment t, Use \n as a delimiter, and the timeline is t1,t2,... ,tN.

Story:
{(*@\color{blue}{story}@*)}

Only output the story with the added timeline, do not provide explanations.
\end{lstlisting}

\begin{lstlisting}
(*@\textbf{Temporal Belief State Chain Construction}@*): 
The following is a sequence of events with a timeline about some characters, that takes place in multiple locations.
Your job is to output only the events on the timeline that character {(*@\color{blue}{character}@*)} can aware of.

Here are a few commonsense rules:
1. If a character is in a certain room/location, they will be aware of all other events happening in that room. This includes other characters entering or leaving the location,  the locations of objects within it, and whether someone has moved an object to another location. 
2. If a character leaves a location and is no longer there, they will no longer be aware of any events occurring at that location. However, they can re-enter the location.
3. A character is aware of all the events that they do.

Story:
{(*@\color{blue}{story}@*)}

What events on the timeline does {(*@\color{blue}{character}@*)} aware of? Only output the events according to the above rules, do not provide an explanation.
\end{lstlisting}

\begin{lstlisting}
(*@\textbf{Time-Aware Belief Question Answer with Belief Compression (First-order ToM questions)}@*): 
Belief Compression: The following is information from the perspective of the character, {(*@\color{blue}{character}@*)}.

Perspective:
{(*@\color{blue}{perspective}@*)}

Output the remaining perspective information after removing the events of characters enter or leave/exit the room/location, do not provide an explanation.

Time-Aware Belief Question Answer: 
{(*@\color{blue}{perspective2}@*)}
You are {(*@\color{blue}{name}@*)}.
Based on the above information, answer the following question:
{(*@\color{blue}{question}@*)}
Keep your answer concise, one sentence is enough. You must choose one of the above choices.
\end{lstlisting}

\begin{lstlisting}
(*@\textbf{Time-Aware Belief Question Answer without Belief Compression (Higher-order ToM questions)}@*): 
{(*@\color{blue}{perspective}@*)}
You are {(*@\color{blue}{name}@*)}.
Based on the above information, answer the following question:
{(*@\color{blue}{question}@*)}
Keep your answer concise, one sentence is enough. You must choose one of the above choices.
\end{lstlisting}

\begin{lstlisting}
(*@\textbf{Time-Aware Belief Solver}@*): 
Perspective1: {(*@\color{blue}{perspective}@*)}
You are {(*@\color{blue}{name}@*)}.
Based on the above information, answer the following question:
{(*@\color{blue}{question}@*)}
Answer1:{(*@\color{blue}{answer}@*)}
Feedback Perspective2: The event corresponding to the period of belief communication between characters {(*@\color{blue}{questionSubject}@*)} and {(*@\color{blue}{questionObject}@*)}: {(*@\color{blue}{common\_belief}@*)} Based on this information, the answer we get to the question:{(*@\color{blue}{question}@*)} is Answer2: {(*@\color{blue}{answer2}@*)}
Consider Perspective1, Feedback Perspective2 and their answers, answer the question: {(*@\color{blue}{question}@*)} again. Keep your answer concise, one sentence is enough. You must choose onea of the above choices.
\end{lstlisting}




\section{Experiments}

\subsection{The Effect of Constructing Temporal Space}

\label{appendix_effect}
\begin{table}[H]
    \centering
    \small
    \setlength\tabcolsep{3.5pt}
    \begin{tabular}{c|cccc}
    \toprule
    \diagbox{ToMI}{Model} & L-7b & w/t & L-13b & w/t\\
    \midrule
    Total & 44.50 & 58.80 \tiny{$\uparrow$14.30} & 51.00 & 60.90\tiny{$\uparrow$9.90}\\
    True-Belief & 50.75 & \textbf{73.00 \tiny{$\uparrow$22.25}} & 50.25 & 60.00\tiny{$\uparrow$9.75}\\
    False-Belief & 28.25 & 30.00 \tiny{$\uparrow$1.75}& 39.25 & 52.00\tiny{$\uparrow$12.75}\\
    Mem-Real & 64.50 & \textbf{88.00 \tiny{$\uparrow$23.50}}& 76.00 & 83.50 \tiny{$\uparrow$7.50}\\
    First-Order & 39.00 & 52.75 \tiny{$\uparrow$13.75}& 54.75 &58.50\tiny{$\uparrow$3.75}\\
    Second-Order & 40.00 & 50.25 \tiny{$\uparrow$10.25}& 34.75 & 52.00\tiny{$\uparrow$17.25}\\
    \bottomrule
    \end{tabular}
    \caption{Performance comparison of the Llama2 series models in 0-shot and 0-shot with timeline settings under ToMI benchmark.}
    \label{tab:accuracy}
\end{table}


\subsection{Full Results}

In Table \ref{tab:10} and \ref{tab:20}, we present the full results of \textsc{TimeToM}\xspace on the ToMI, BigToM, and FanToM benchmarks.
\label{appendix3}
{\renewcommand{\arraystretch}{1}
    \begin{table*}[t!] \begin{center}
    \begin{adjustbox}{width=\linewidth}
    \begin{tabular}{lcccccccc}
            \toprule
            \multirow{2}{*}{Model}  
            & \multicolumn{6}{c}{\makecell{ToMI}}  
            & \multicolumn{2}{c}{\makecell{BigTOM}}\\
            \cmidrule(r{0.3em}){2-7} \cmidrule(r{0.3em}){8-9}  
            & Total     & True-Belief   & False-Belief  &Mem-Real & First-Order   & Second-Order    & True-Belief      & False-Belief \\
            \midrule
            \multicolumn{9}{l}{\cellcolor{lightgray}0-Shot} \\
            Llama2-7b-chat & 44.50   &  50.75   & 28.25    &  64.50   &  39.00  &  40.00  & 51.50  & 53.50\\
            Llama2-13b-chat &  51.00  &  50.25   & 39.25    &  76.00   &  54.75  &  34.75  & 64.00  & 46.50\\
            GPT-3.5-turbo &  68.60  &  54.25   &  67.25 &100.00  &  68.75   &  52.75  &  87.50  & 69.50   \\
            GPT-4 & 66.50   &  90.75   &  25.50 &100.00  &  50.75   &  65.50  &  96.00  &  99.00  \\

            \multicolumn{9}{l}{\cellcolor{lightgray}0-Shot-CoT} \\
            Llama2-7b-chat & 43.70   &  58.75   &  24.00   &  53.00   &  45.00  & 37.75 & 61.50  & 39.50   \\
            Llama2-13b-chat &  45.00  &  63.50   &  16.50   &   65.00  & 43.00   &  37.00  &  62.00 & 52.50 \\
            GPT-3.5-turbo &  64.10  &  77.50   & 34.00  & 97.50  &  58.50   &  53.00  &   90.00 & 71.50   \\
            GPT-4 &  74.40  &  61.75   &  74.25 &100.00  &  73.75   &  62.25  &  96.50  &  99.00  \\

            \multicolumn{9}{l}{\cellcolor{lightgray}\textsc{SimToM}\xspace} \\
            Llama2-7b-chat &  48.10  &   46.50  &   40.00 & 67.50 &   47.25  &  39.25  & 37.50   &  75.00  \\
            Llama2-13b-chat &  61.10  &  72.00   &  35.50 &90.50  &  53.75   &  53.75  &  53.00  &  62.50  \\
            GPT-3.5-turbo &  72.80  &  51.00   &  81.00 & 100.00  &  74.75   &  57.25  &  90.00  &   78.00 \\
            GPT-4 & 87.80   &  81.75   &  87.75 & 100.00  &   93.75  &  75.75  &  94.00  &  98.00  \\

            \multicolumn{9}{l}{\cellcolor{lightgray}\textbf{\textsc{TimeToM}\xspace}} \\
            \multirow{2}{*}{Llama2-7b-chat} &  64.30  &  67.00   &   47.25 & 93.00  &   56.50  &  57.75  &  53.00  & 84.50   \\ & \textbf{(+19.80,+16.20)} & \textbf{(+16.25, +20.50)} & \textbf{(+19.00, +7.25)} & \textbf{(+28.50, +25.50)} & \textbf{(+17.50, +9.25)} & \textbf{(+17.75, +18.50)} & \textbf{(+1.50, +15.50)} & \textbf{(+31.00, +9.50)}\\
            \multirow{2}{*}{Llama2-13b-chat} &  67.20  &  73.50   & 44.75 &99.50   &  61.25   &  57.00  &  66.00  & 89.50   \\ 
            & \textbf{(+16.20, +6.10)} & \textbf{(+23.25, +1.50)} & \textbf{(+5.50, +9.25)} & \textbf{(+23.50, +9.00)} & \textbf{(+6.50, +7.50)} & \textbf{(+22.25, +3.25)} & \textbf{(+2.00, +13.00)} & \textbf{(+43.00, +27.00)}\\
            \multirow{2}{*}{GPT-3.5-turbo} & 80.80   &  70.00   &  82.00 & 100.00  &  80.50   &  71.50  &  91.50  &  96.00  \\
            & \textbf{(+12.20, +8.00)} & \textbf{(+15.75, +19.00)} & \textbf{(+14.75, +1.00)} & \textbf{(+0.00, +0.00)} & \textbf{(+11.75, +5.75)} & \textbf{(+18.75, +14.25)} & \textbf{(+4.00, +1.50)} & \textbf{(+26.50, +18.00)}\\
            \multirow{2}{*}{GPT-4} &  96.00  &  90.75   &  98.75 & 100.00  &  95.50   &  94.50  & 95.00   &  99.00  \\
            & \textbf{(+29.50, +8.20)} & \textbf{(+0.00, +9.00)} & \textbf{(+73.25, +11.00)} & \textbf{(+0.00, +0.00)} & \textbf{(+44.75, +1.75)} & \textbf{(+29.00, +18.75)} & \textbf{(-1.00, +1.00)} & \textbf{(+0.00, +1.00)}\\
            \bottomrule
    \end{tabular}
    \end{adjustbox}
    \caption{
          The full results of \textsc{TimeToM}\xspace on the ToMI, BigToM benchmarks. Mem-Real can be viewed as zeroth-order ToM question.
    }
    \vspace{-10pt}
    \label{tab:10}
\end{center}\end{table*}}

{\renewcommand{\arraystretch}{1}
    \begin{table*}[t!] \begin{center}
    \begin{adjustbox}{width=\linewidth}
    \begin{tabular}{lccccccccccc}
            \toprule
            \multirow{2}{*}{Model}  
            & \multicolumn{1}{c}{\multirow{2}{*}{\makecell{{\textcolor[RGB]{0,201,87} {\textbf{\textsc{All*}}}}\\Question\\Types}}}
            & \multicolumn{4}{c}{\makecell{{Belief}\\{Questions}}}  
            & \multicolumn{3}{c}{\makecell{{Answerability}\\{Questions}}}
            & \multicolumn{3}{c}{\makecell{{Infoaccess}\\{Questions}}}\\
            \cmidrule(r{0.3em}){3-6} \cmidrule(r{0.3em}){7-9}  \cmidrule(r{0.3em}){10-12}  
            &     & Overall     & First-order   & Third-acyc   & Third-cyc   & {\textcolor[RGB]{0,201,87} {\textbf{All}}}    & List      & Binary  & {\textcolor[RGB]{0,201,87} {\textbf{All}}}    & List      & Binary\\
            \midrule
            \multicolumn{12}{l}{\cellcolor{lightgray}0-Shot} \\
            Llama2-70b-chat & 0.0 & 6.5 & 8.7 & 0.0 & 5.7 & 4.3 & 30.4 & 60.4 & 8.7 & 21.7 & 75.4\\
            GPT-4 & 8.7 & 76.2 & 73.0 & 77.1 & 85.7 & 23.5 & 44.3 & 73.8 & 23.5 & 28.7 & 90.3\\
            \multicolumn{12}{l}{\cellcolor{lightgray}0-Shot-CoT} \\
            Llama2-70b-chat & 3.5 & 69.7 & 64.3 & 77.1 & 80.0 & 11.3 & 45.2 & 66.8 & 13.9 & 47.0 & 72.8\\
            GPT-4 & 10.4 & 75.1 & 73.0 & 74.3 & 82.9 & 25.2 & 48.7 & 75.6 & 34.8 & 47.8 & 89.4\\
            \multicolumn{12}{l}{\cellcolor{lightgray}\textsc{TimeToM}\xspace} \\
            \multirow{3}{*}{Llama2-70b-chat} & 6.1 & 79.0 & 75.7 & 80.0 & 88.6  & 17.4 & 51.3 & 69.0 & 15.7 & 60.0 & 68.2\\ 
            & {\textcolor[RGB]{0,201,87} {\textbf{(+6.1, +2.6)}}} & \textbf{(+72.5, +9.3)} & \textbf{(+67.0, +11.4)} & \textbf{(+80.0, +2.9)} & \textbf{(+82.9, +8.6)} & {\textcolor[RGB]{0,201,87} {\textbf{(+13.1, +6.1)}}} & \textbf{(+20.9, +6.1)} & \textbf{(+8.6, +2.2)} & {\textcolor[RGB]{0,201,87} {\textbf{(+7.0, +1.8)}}} & \textbf{(+38.3, +13.0)} & \textbf{(-7.2, -4.6)}\\
            & {\textcolor[RGB]{0,201,87} {\textbf{($\times$ $\infty$, $\times$ 1.7)}}} & & & & & {\textcolor[RGB]{0,201,87} {\textbf{($\times$ 4.0, $\times$ 1.5)}}}& & &{\textcolor[RGB]{0,201,87} {\textbf{($\times$ 1.8, $\times$ 1.1)}}} & &\\
            \multirow{3}{*}{GPT-4} & 41.7 & 93.0 & 93.1 & 94.3 & 91.5 & 51.3 & 62.6 & 90.7 & 52.2 & 63.5 & 92.0\\
            & {\textcolor[RGB]{0,201,87} {\textbf{(+33.0, +31.3)}}} & \textbf{(+16.8, +17.9)} & \textbf{(+20.1, +20.1)} & \textbf{(+17.2, +20.0)} & \textbf{(+5.8, +8.6)} & {\textcolor[RGB]{0,201,87} {\textbf{(+27.8, +26.1)}}} & \textbf{(+18.3, +13.9)} & \textbf{(+16.9, +15.1)} & {\textcolor[RGB]{0,201,87} {\textbf{(+28.7, +17.3)}}} & \textbf{(+34.8, +15.7)} & \textbf{(+1.7, +2.6)}\\
            & {\textcolor[RGB]{0,201,87} {\textbf{($\times$ 4.8, $\times$ 4.0)}}} & & & & & {\textcolor[RGB]{0,201,87} {\textbf{($\times$ 2.2, $\times$ 2.0)}}}& & &{\textcolor[RGB]{0,201,87} {\textbf{($\times$ 2.2, $\times$ 1.5)}}} & &\\



            \bottomrule
    \end{tabular}
    \end{adjustbox}
    \caption{The full results of \textsc{TimeToM}\xspace on the FanToM benchmark.
    }
    \vspace{-10pt}
    \label{tab:20}
\end{center}\end{table*}}

\end{document}